%% file: main.tex
\documentclass[10pt,twocolumn,letterpaper]{article}

\usepackage{iccv}              % To produce the CAMERA-READY version
\input{preamble}
\input{math_commands}

\definecolor{iccvblue}{rgb}{0.21,0.49,0.74}
\usepackage[pagebackref,breaklinks,colorlinks,allcolors=iccvblue]{hyperref}

\def\paperID{3654} % *** Enter the Paper ID here
\def\confName{ICCV}
\def\confYear{2025}

\title{ViT-Linearizer: Distilling Quadratic Knowledge into Linear-Time Vision Models}

\author{Guoyizhe Wei\\
Johns Hopkins University\\
Baltimore, US\\
\and
Rama Chellappa\\
Johns Hopkins University\\
Baltimore, US\\
}

\begin{document}
\maketitle
\input{sec/0_abstract}    
\input{sec/1_intro}
\input{sec/2_related_work}
\input{sec/3_method}
\input{sec/4_experiment}
\input{sec/5_conclusion}
\input{sec/acknowledge}

{
    \small
    \bibliographystyle{ieeenat_fullname}
    \bibliography{main}
}
\input{sec/X_Appendix}

\end{document}

%% file: preamble.tex
\usepackage{algorithm}

\usepackage{graphicx}
\usepackage{url}
\usepackage{multirow}
\usepackage{makecell}
\usepackage{booktabs}
\usepackage{subcaption}
\usepackage{xcolor}
\usepackage{colortbl}
\usepackage{float}
\usepackage{stfloats}
\usepackage{pifont}
\usepackage{natbib}
\usepackage{etoolbox}

\usepackage{amsmath}
\usepackage{siunitx} % For aligning numbers at the decimal point
\usepackage{algpseudocode}
\definecolor{citecolor}{RGB}{34,139,34}
\definecolor{demphcolor}{gray}{.5}
\definecolor{Graylight}{gray}{0.9}
\newcommand{\demph}[1]{\textcolor{demphcolor}{#1}}
\newcommand{\gtext}[1]{\textcolor{citecolor}{#1}}

\newlength\savewidth\newcommand\shline{\noalign{\global\savewidth\arrayrulewidth
  \global\arrayrulewidth 1pt}\hline\noalign{\global\arrayrulewidth\savewidth}}
\newcommand{\tablestyle}[2]{\setlength{\tabcolsep}{#1}\renewcommand{\arraystretch}{#2}\centering\footnotesize}
\renewcommand{\paragraph}[1]{\vspace{1.25mm}\noindent\textbf{#1}}

\usepackage{pifont}
\newcommand{\cmark}{\ding{51}}
\newcommand{\xmark}{\ding{55}}

%% file: math_commands.tex
\usepackage{amsmath,amsfonts,bm}

\def\eqref#1{equation~\ref{#1}}
\def\1{\bm{1}}

\def\vf{{\bm{f}}}

\def\vk{{\bm{k}}}

\def\vq{{\bm{q}}}

\def\vv{{\bm{v}}}

\def\vx{{\bm{x}}}
\def\vy{{\bm{y}}}

\def\mA{{\bm{A}}}

\def\mS{{\bm{S}}}

\def\mW{{\bm{W}}}

\def\mY{{\bm{Y}}}

\DeclareMathAlphabet{\mathsfit}{\encodingdefault}{\sfdefault}{m}{sl}
\SetMathAlphabet{\mathsfit}{bold}{\encodingdefault}{\sfdefault}{bx}{n}

\def\sR{{\mathbb{R}}}

%% file: sec/0_abstract.tex
\begin{abstract}

Vision Transformers (ViTs) have delivered remarkable performence through global self-attention, yet their quadratic complexity can become prohibitive for high-resolution inputs. In this work, we present ViT-Linearizer, a cross-architecture distillation framework that transfers rich ViT representations into a linear-time, recurrent-style model. Our approach leverages 1) activation matching, an intermediate constraint that encourages a student to align its token-wise dependencies with those produced by the teacher, and 2) masked prediction, a contextual reconstruction objective that requires the student to predict the teacher’s representations for unseen (masked) tokens, to effectively distill the quadratic self-attention knowledge into the student while maintaining efficient complexity. Empirically, our method provides notable speedups particularly for high-resolution tasks, significantly addressing the hardware challenges in inference. Additionally, it also elevates Mamba-based architectures’ performance on standard vision benchmarks, achieving a competitive 84.3\% top-1 accuracy on ImageNet with a base-sized model. Our results underscore the good potential of RNN-based solutions for large-scale visual tasks, bridging the gap between theoretical efficiency and real-world effectiveness.

\end{abstract}

%% file: sec/1_intro.tex
\section{Introduction}
\label{sec:intro}

\begin{figure}
    \centering
    \includegraphics[width=\linewidth]{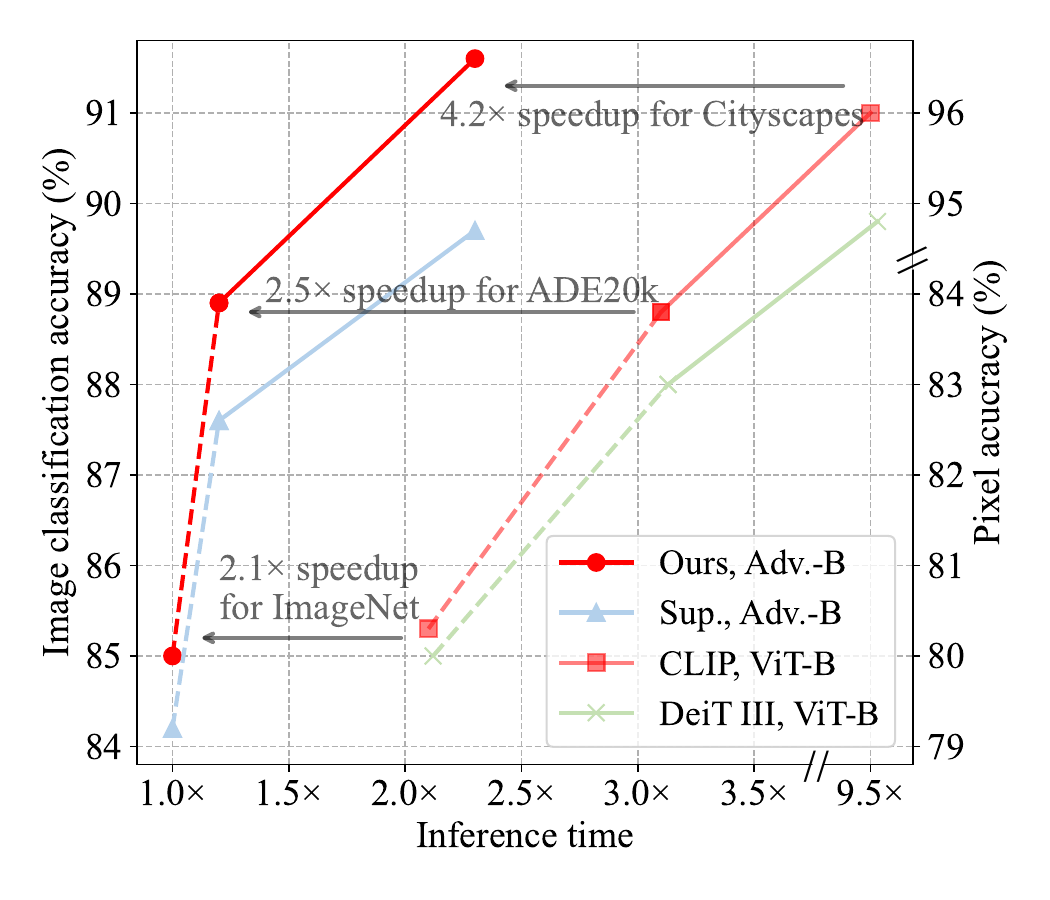}
    \vspace{-1cm}
    \caption{\textbf{Accuracy-efficiency trade-offs.} We distill CLIP's ViT-Base model into a linear-time Adventurer-Base~\cite{adventurer} (with Mamba-2~\cite{mamba2} token mixers), which exhibits substantially superior accuracy-efficiency trade-offs across various datasets and tasks.}
    \label{fig:perform}
\end{figure}

Recent advances in training foundational Vision Transformers (ViTs)~\cite{vit} have led to significant breakthroughs in visual representation learning, with numerous self-attention-based architectures now dominating the applications in visual understanding, generation, and multi-modal reasoning tasks~\cite{deit,mae,dino,stablediff,dit,clip,llama,llava,gpt4}. As self-attention requires each token to compute its correspondence with every other positions in the sequence, ViT models are able to produce robust token-level (\ie, patch-level) representations even in weakly supervised or unsupervised training scenarios~\cite{vit,dino,dinov2,register,clip,maskclip,sclip,defo}. Nonetheless, this strong local representation capability comes with a quadratic computation complexity with respect to sequence length, raising a considerable challenge in inference efficiency and hardware capacities when learning from long contexts. It is noteworthy that this computational overhead was not particularly prohibitive in standard benchmarks featuring medium-resolution and coarse-granularity visual tasks that were historically focused on~\cite{imagenet,voc12,pascalcontext,ade20k,coco,vqav2}. However, as the demand for processing high-resolution, high-fidelity visual inputs rapidly grows, the question of how to efficiently leverage the \textbf{\textit{quadratic knowledge}} learned by ViTs has become increasingly important.

As alternatives, RNN-fashion token mixers such as Mamba~\cite{mamba,mamba2}, RWKV~\cite{rwkv}, and xLSTM~\cite{xlstm} have recently been introduced in vision tasks, showcasing competitive prediction results and superior accuracy-computation trade-offs compared with ViTs~\cite{vim,adventurer,visionrwkv,visionlstm}. These recurrent vision models, whose computational cost and memory requirements scale linearly with sequence length, emerge as a potential solution to address the computationally explosive issues posed by the quadratic complexity of self-attention when processing long contexts. However, unlike ViTs—into which substantial effort and resources have already been invested—the exploration of recurrent vision models has so far been restricted to relatively small data scales and modest model sizes. These limitations motivate us to develop a cross-architecture distillation approach that can effectively transfer the capabilities of ViTs into the linear-time recurrent models such as Vision Mamba~\cite{vim} and Adventurer~\cite{adventurer}.

Through extensive investigations into cross-architecture transfer strategies, we identify that a naive distillation between ViTs and Mamba fails to produce strong student models; instead, the key to transferring quadratic knowledge with minimal loss to linear-time recurrent models lies in \textbf{\textit{activation matching}} and \textbf{\textit{masked prediction}}. Specifically, we find that compared to the final-layer outputs, ViT models typically capture more informative content in their intermediate activation maps (or, in an analogous formulation, attention maps). These activation maps directly reflect the token-wise dependencies learned under self-attention’s quadratic computational cost, which play a critical role in ViTs’ robust representational capabilities. Inspired by this observation, we introduce an activation matching constraint at multiple intermediate layers which is implemented as an $\ell_2$ loss that minimizes the distance between the normalized activation scores of the teacher and student models. Intuitively, this mechanism enables the new model to learn more precise local representations and encourages the recurrent students to behave similarly to their ViT teachers. Empirically, activation matching leads to substantial gains in predictive performance (\eg, 1.2\% accuracy improvement on ImageNet) and effectively addresses the noisy activation issues in recurrent vision models (see Figure~\ref{fig:attn}).

Moreover, similar to prior observations in ViT pretraining, where masked distillation was found to facilitate robust feature representations and potentially provides students with reasoning capabilities exceeding those of their teachers~\cite{ibot,maskdistill,dinov2}, we find that the cross-architecture distillation between ViT and Mamba also derives substantial benefit from the masked prediction strategy. Specifically, for the final-layer features, we do not directly distill them from the teacher model but instead employ a masking strategy that requires the student model to predict the visual representations of unseen tokens---following the prior practice of masked image modeling~\cite{beit,mae,ibot,maskdistill}, we mask out a portion of the image patches in the student model and replace them with a learnable [mask] token, and then optimize the student by aligning the [mask] token outputs with the teacher’s outputs at the corresponding positions. 

We term this new approach \textit{\textbf{ViT-Linearizer}}. Overall, the ViT-Linearizer confers two primary benefits that serve as the core contributions of this work. First, the ViT-Linearizer effectively transfers the knowledge learned by ViTs under their quadratic complexity costs into linear-time recurrent vision models with minimal compromise in predictive performance, providing a promising solution to mitigating ViTs’ considerable inference cost and elevated memory demands. In our experiments, we successfully distill the ViT-Base encoder from CLIP~\cite{clip} into a linear-complexity Adventurer-Base model~\cite{adventurer}, which ultimately achieves a highly competitive fine-tuning accuracy of 84.3\% on ImageNet. As shown in Figure~\ref{fig:perform}, when processing high-resolution inputs, our distilled model offers significant inference speedup (\eg, 4.2$\times$ for Cityscapes~\cite{cityscapes} semantic segmentation) compared to its ViT teacher. Moreover, from the perspective of improving the practical performance of recurrent vision models, the ViT-Linearizer produces a new state of the art for the Mamba-based architecture on various vision tasks. Remarkably, we successfully boost  Adventurer-Large’s test accuracy on the standard ImageNet-1k classification benchmark from 83.4\% to 85.0\%, demonstrating that given robust supervision, linear-complexity recurrent vision models can also achieve prediction performance on par with ViTs.

We hope this work offers fresh insights into efficient inference for visual foundation models. In general, we rely on large-scale, high-complexity models during pretraining to thoroughly exploit the statistical knowledge in the training data, while aiming to reduce both compute time and hardware requirements during inference. Compared with model pruning or distillation into smaller models---approaches that often suffer considerable performance degradation and fail to fundamentally address the high complexity of self-attention---our cross-architecture distillation method enables efficient inference over long sequences with a far smaller accuracy sacrifice. We envision this approach leading to a new transfer learning paradigm, where reducing model complexity to boost inference efficiency in downstream tasks allows models to seamlessly inherit pretraining knowledge while more effectively adapting to long-context representations.

%% file: sec/2_related_work.tex
\section{Related Work}
\paragraph{Linear-complexity token mixers.} Overcoming the quadratic overhead in full self-attention has motivated various approaches that constrain or approximate the attention mechanism to achieve linear or near-linear complexity. Notable examples include Linear attention methods that replace the softmax with kernel-based or low-rank approximations~\cite{linattn,linformer,performer,nystromformer}, as well as recurrent or token-by-token architectures exemplified by Mamba~\cite{mamba,mamba2} and RWKV~\cite{rwkv}. In parallel, improved versions of classical recurrent models, such as the LSTM \cite{lstm} and its variants~\cite{lstm01,xlstm,retnet}, have been explored to capture long-range dependencies in a more memory-efficient manner. These approaches fundamentally trade the global pairwise interactions of full attention for efficient, linear-scale operations, making them appealing for large-resolution or long-sequence tasks in both language and vision domains.

\paragraph{RNN-fashion vision models.} While most modern vision architectures rely on self-attention or convolutional designs, there is a rising interest in recurrent approaches as they combine the advantages of the global receptive field and linear or near-linear time and space complexity. One of the primary architectural designs simply takes the macro structure of ViT as a blueprint and replaces self-attention by recurrent modules~\cite{vim,mambar,visionlstm,visionrwkv,videomamba,plainmamba,adventurer,mvar,arm}, complemented by additional designs that enable the otherwise unidirectional token mixer to operate smoothly in typical bidirectional vision tasks~\cite{vim,adventurer}. In addition, other designs adopt hierarchical or hybrid structures to further enrich the model's expressive capacity, often by incorporating multi-scale or progressive token downsampling so as to capture both fine-grained details and broader contextual dependencies in a unified framework~\cite{vmamba,mambavision}. These recurrent-style vision models have recently attracted significant attention due to their intrinsic efficiency benefits, and have advanced rapidly with broad applications in both multi-modal understanding~\cite{vlmamba,cobra} and generation tasks~\cite{zigma,dig,dim}.

\paragraph{Cross-architecture distillation.} Cross-architecture distillation offers a particularly efficient way to transfer pretrained knowledge. Compared to training from scratch that typically requires considerable resources and time, this approach allows newly proposed architectures to quickly assess their practical performance across various tasks at an early exploration stage. Notably, in the early stages of ViT research, DeiT~\cite{deit} demonstrated how a well-pretrained CNN could be distilled into a ViT to achieve competitive performance, sparking numerous follow-up studies on effectively applying Transformer architectures in vision. More relevantly, Transformer-to-RNN distillation has recently garnered significant attention, with numerous pioneering efforts already demonstrating its effectiveness and feasibility in language and vision-language tasks~\cite{distillquadra,kasai2021finetuning,zhang2024hedgehog,zhang2024lolcats,liao2025multimodal,wang2024mamba,mercat2024linearizing,flowar}. This work aims to explore how to advance self-attention linearization from the perspective of purely vision-driven tasks, proposing a cross-architecture distillation framework that transfers the strong representational capabilities of pretrained Transformers into efficient recurrent models, thereby bridging the gap between theoretical efficiency and practical performance in the visual domain.

%% file: sec/3_method.tex
\section{Method}

\begin{figure*}[t]
    \centering
    \includegraphics[width=0.95\textwidth]{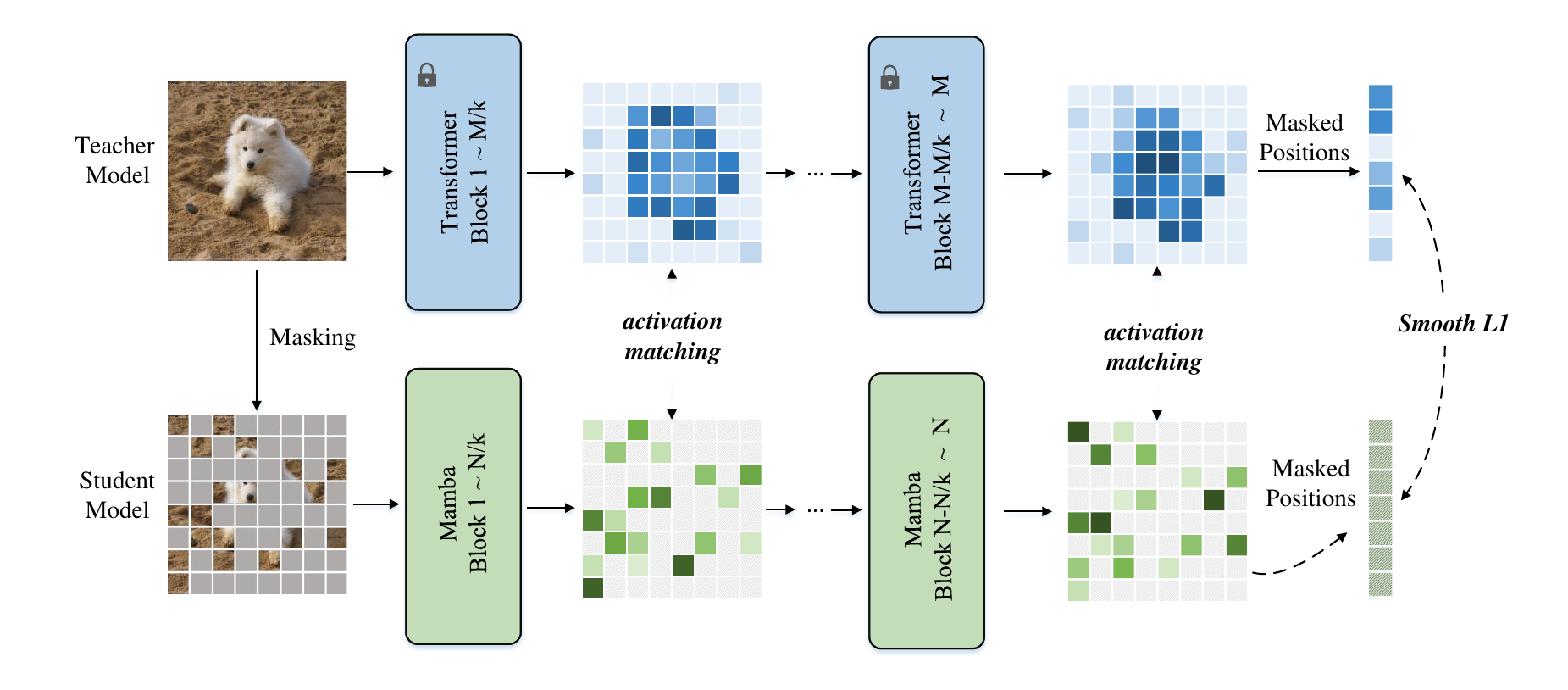}
    \caption{\textbf{Overview of our cross-architecture distillation pipeline.} We feed the complete input image to the frozen teacher (ViT) while providing a randomly masked image to the student (a recurrent model such as Adventurer~\cite{adventurer}). At $K$ intermediate stages, we enforce a token-wise matching between the teacher’s and student’s activation maps. In the final layer, the student predicts the teacher’s representations for the unseen (masked) tokens. Only the student network is trained, while the teacher remains frozen throughout.}
    \label{fig:vis}
\end{figure*}

\subsection{Problem Formulation}
Overall, our primary goal is to distill the knowledge learned by a well-pretrained ViT---which operates with a quadratic time complexity in self-attention---into a linear-complexity recurrent model for efficient visual inference. Concretely, we employ the Adventurer~\cite{adventurer} architecture as our student, which is equipped with the latest Mamba-2~\cite{mamba2} token mixer and has been carefully optimized to achieve fast inference speed while retaining competitive accuracy in visual tasks. Like a typical ViT, Adventurer has a plain architecture that first divides an input image into non-overlapping patches and processes them through alternating token mixer and channel mixer modules. Formally, self-attention
\begin{equation}
\begin{aligned}
    \vq, \vk, \vv = \vx\mW_{Q}, \vx\mW_{K}, \vx\mW_{V}, \\
    \vy=\mbox{softmax}(\vq\vk^T/\sqrt{d})\vv,
\end{aligned}
\end{equation}
forces each token to compute dependencies to every other positions, incurring $\mathcal{O}(L^2)$ complexity for a sequence of length $L$. Here, the $\vx$ and $\vy$ denote input and output, and $\mW_{Q},\mW_{K},\mW_{V}\in\sR^{d\times d}$ are learnable projection matrices. By contrast, the Mamba-2 token mixer operates with
\begin{equation}
\begin{aligned}
    \vq, \vk, \vv, \delta = \vx\mW_{Q}, \vx\mW_{K}, \vx\mW_{V}, \vx\mW_{\delta}, \\
    \vy = \left(\exp(-\mbox{softplus}(\delta)\alpha)\mS + \vv\vk^T\right)\vq,
\end{aligned}
\end{equation}
where $\mW_{Q},\mW_{K},\mW_{V}\in\sR^{d\times d}, \mW_\delta\in\sR^{d\times1}, \alpha\in\sR$ are trainable parameters and $\mS$ denotes the hidden state calculated in a recurrent formulation:
\begin{equation}
    \mS_t = \exp(-\mbox{softplus}(\delta_t)\alpha)\mS_{t-1} + \vv_t\vk_t^T,
\end{equation}
where $t$ denotes the token index in the sequence.

Similar to the derivation in prior studies~\cite{mamba,mamba2,vmamba,mambalinearattention}, here we can also observe that self-attention and Mamba-2 share substantial similarities in their overall formulations, whereas Mamba-2 employs a simplified strategy for efficient inference---it accumulates token-wise dependencies in a stateful fashion and brings the complexity down to $\mathcal{O}(L)$. Given this intrinsic similarity, our proposed ViT-Linearizer method thus focuses on bridging the representational gap between these two token mixers by distillation. That is, we aim to transfer the knowledge encoded in the self-attention maps of the teacher ViT—learned at high computational cost—into the linear-time Mamba-2 formulation within Adventurer. This allows the distilled model to inherit strong token-wise representations while substantially reducing the inference overhead, thereby facilitating downstream tasks that demand efficient yet high-performing vision models.

\subsection{Activation Matching}
As discussed in Section~\ref{sec:intro}, a central insight of our distillation approach is that ViTs owe much of their representational capabilities to the token-to-token dependencies learned through self-attention, which cannot be simply transferred by the naive distillation approach~\cite{distillation}. We introduce the activation matching mechanism to enable effective knowledge transfer of token-wise correspondences from ViT to recurrent models. Formally, we suppose the teacher (a ViT) and student (an Adventurer with Mamba-2) models have $M$ and $N$ basic blocks, respectively, and partition the blocks in both models into $K$ stages. For example, for the base-sized ViT and Adventurer models, we have $M=N=12$ and set $K=4$ in our default configuration. At each stage, we have an $\sR^{L\times d}$ feature map where $L$ denotes the sequence length and $d$ is the dimensionality of latent features. We then compute \textit{\textbf{activation maps}} for both teacher ($\mA_{\text{tea}}^k\in\sR^{L\times L}$) and student ($\mA_{\text{stu}}^k\in\sR^{L\times L}$) by taking the pairwise cosine similarities among all tokens:
\begin{equation}\label{eq:act}
    \mA_{\text{tea}}^k=\frac{\vf_{\text{tea}}^k(i) \cdot \vf_{\text{tea}}^k(j)}{||\vf_{\text{tea}}^k(i)||_2||\vf_{\text{tea}}^k(j)||_2}, \mA_{\text{stu}}^k=\frac{\vf_{\text{stu}}^k(i) \cdot \vf_{\text{stu}}^k(j)}{||\vf_{\text{stu}}^k(i)||_2||\vf_{\text{stu}}^k(j)||_2},
\end{equation}
for $i,j=1,\dots,L$ and stage $k=1,\dots,K$. Here, $\vf_{\text{tea}}^k(\cdot)$, $\vf_{\text{stu}}^k(\cdot)\in\sR^d$ denote the teacher’s and student’s token embeddings at stage $k$. Note that for easy notations, we simply assume the teacher and student models share the same latent dimension, yet different $d$ between teacher and student does not affect the calculation of activation maps.

We next normalize each row of $\mA_{\text{tea}}^k$ and $\mA_{\text{stu}}^k$ via $\ell_2$ norm, and define the activation matching loss as:
\begin{equation}
    \mathcal{L}_\text{act}= \frac{1}{KL}\sum_{k=1}^K\sum_{i=1}^L\left[ 1 - <\overline{\mA}_{\text{tea}}^k(i, :), \overline{\mA}_{\text{stu}}^k(i, :)> \right],
\end{equation}
where $\overline{\mA}_{\text{tea}}^k(i, :)$ and $\overline{\mA}_{\text{stu}}^k(i, :)$ are the normalized $i$-th rows of the teacher and student activation maps at stage $k$. Notably, this procedure closely parallels the token-wise correspondence computation in self-attention, and the loss function $\mathcal{L}_\text{act}$ itself involves $\mathcal{O}(L^2)$ computations---we term this loss function a quadratic constraint and will further empirically demonstrate that \textbf{\textit{the quadratic constraint is a necessary component for quadratic knowledge distillation.}}

\subsection{Masked Prediction}\label{sec:mask}
Similar to earlier observations in pretraining ViTs~\cite{mae,beit,maskfeat,ibot,maskdistill}, we find that masked prediction, instead of aligning the entire feature map with the supervision information, makes a substantial difference for the representational capability of Mamba-based architectures. Specifically, we follow the standard asymmetric structure~\cite{beit,ibot,maskdistill} to feed the teacher model with the entire image and the student model with masked input which is implemented by randomly replacing a portion of patch tokens with a learnable [mask] token. The teacher and student models finally produce outputs $\mY_\text{tea}$, $\mY_\text{stu}\in\sR^{L\times d}$. We discard the unmasked positions and compute the masked prediction loss by
\begin{equation}
    \mathcal{L}_\text{mask}=\frac{1}{aL}\sum_{i\in\Omega}\text{Smooth}\ell_1(\mY_\text{tea}(i,:), \mY_\text{stu}(i,:)),
\end{equation}
where $a\in[0, 1]$ denotes the mask ratio and $\Omega$ is the set of unmasked indices. We choose Smooth$\ell_1$ as the basic loss function since it facilitates robust convergence and yields slightly better predictive performance than $\ell_2$ losses. In our experiments, we simply follow MAE~\cite{mae} to set $a=0.75$. Again, for ease of notation, here we assume teacher and student always have the same latent dimension. In the actual implementation, we additional apply a linear projection layer to keep their features in the same latent space.

\paragraph{Integrating activation matching with masks.} It is noteworthy that the masking mechanism changes the effective representation space of the student---for the masked positions, the intermediate features are predictions of unseen information, rather than the representations of corresponding input tokens like those in the teacher model. Moreover, applying activation matching to unseen tokens leads to direct information leakage, causing the masked prediction in the final layer to easily collapse. To enable compatible operations for both mechanisms, we restrict the scope of activation matching to only those tokens visible to the student, meaning that, in calculating the activation maps, only unmasked tokens are considered in Equation~\ref{eq:act}, which eventually leads to $\sR^{(1-a)L\times(1-a)L}$ sized activations for both $\mA_{\text{tea}}^k$ and $\mA_{\text{stu}}^k$. A detailed ablation of how the two mechanisms affect each other can be found in Table~\ref{tab:component} and Section~\ref{sec:abl}. The total loss function of our ViT-Linearizer can be represented as $\mathcal{L} = \mathcal{L}_\text{act} + \lambda\mathcal{L}_\text{mask}$ and we by default set $\lambda=1$ since we find this coefficient not to be sensitive within a reasonable range (\eg $\lambda\in[0.2, 5.0]$). The overall framework of our ViT-Linearizer approach is shown in Figure~\ref{fig:vis}.

%% file: sec/4_experiment.tex
\section{Experiments}

\subsection{Experimental Setup}

\paragraph{Models.} We utilize the vision branch of CLIP~\cite{clip} (ViT-Base/16) as our default teacher model, leveraging its large-scale, weakly supervised pretraining advancements over diverse domains. Prior work has shown that the CLIP vision encoder outperforms similarly sized pretrained networks on a wide range of downstream tasks~\cite{maskdistill}, making it well-suited for transferring the quadratic knowledge learned via self-attention. In addition, for ablation studies, we experiment with both a supervised pretrained ViT from DeiT-III~\cite{deit3} and an unsupervised pretrained ViT from MAE~\cite{mae}, as well as different ViT model sizes, to illustrate the broad applicability of our ViT-Linearizer approach. For the student model, we employ Adventurer~\cite{adventurer}, a Mamba-based plain architecture recognized for its exceptionally fast inference among recurrent vision models. Specific configurations of teacher and student networks, including layer counts and hidden dimensions, are detailed in Appendix.

\paragraph{Data and Downstream Tasks.} For the distillation phase, we use the standard ImageNet-1k dataset~\cite{imagenet} which contains 1.28 million training images. We do not utilize the class labels during this stage. To evaluate the student models, we fine-tune them on downstream tasks including ImageNet classification, ADE20K~\cite{ade20k} and Cityscapes~\cite{cityscapes} semantic segmentation which features higher input resolutions that raise substantial efficiency challenges for ViT backbones. In detail, the ADE20K dataset comprises 20k images labeled across 150 semantic categories. We crop the inputs into 512$\times$512 pixels for both training and inference. The Cityscapes dataset contains 5,000 finely annotated and high-resolution images for training and validation. In our experiments, we resize the inputs into 512$\times$1024 pixels.

\subsection{Main Results}

\begin{table*}[t]
\centering
\tablestyle{5pt}{1.1}
\begin{tabular}{llc|cccrrr}
    Method & Model & Token mixer & Pt. data & Supervision & Input size & Memory ($\downarrow$) & Throughput ($\uparrow$) & Accuracy(\%) \\\shline
    \textit{\demph{CLIP~\cite{clip}}} & \textit{\demph{ViT-Base/16}} & \textit{\demph{self-attention}} & \textit{\demph{WIT-400M}} & \textit{\demph{text}} & \textit{\demph{224$\times$224}} & \textit{\demph{14.4 GB}} & \textit{\demph{613}} & \textit{\demph{84.7}} \\

    MAE~\cite{mae} & ViT-Base/16 & self-attention & IN1k & pixel & 224$\times$224 & 14.4 GB & 612 & 83.6 \\
    
    supervised & DeiT-Base~\cite{deit} & self-attention & IN1k & class & 224$\times$224 & 14.4 GB & 613 & 81.8 \\
    supervised & DeiT-III-Base~\cite{deit3} & self-attention & IN1k & class & 224$\times$224 & 14.5 GB & 608 & 83.8 \\
    supervised & Vim-Base~\cite{vim} & Mamba~\cite{mamba} & IN1k & class & 224$\times$224 & 20.0 GB & 180 & 81.9 \\
    supervised & Adventurer-Base~\cite{adventurer} & Mamba-2~\cite{mamba2} & IN1k & class & 224$\times$224 & 13.0 GB & 736 & 82.6 \\\rowcolor{orange!10}
    ViT-Linearizer & Adventurer-Base~\cite{adventurer} & Mamba-2~\cite{mamba2} & IN1k & CLIP & 224$\times$224 & 13.0 GB & 736 & \bf 84.3 \\\hline

    \textit{\demph{CLIP~\cite{clip}}} & \textit{\demph{ViT-Base/16}} & \textit{\demph{self-attention}} & \textit{\demph{WIT-400M}} & \textit{\demph{text}} & \textit{\demph{448$\times$448}} & \textit{\demph{$>$80 GB}} & \textit{\demph{95}} & \textit{\demph{85.3}} \\
    supervised & DeiT-Base~\cite{deit} & self-attention & IN1k & class & 448$\times$448 & $>$80 GB & 95 & 83.4 \\
    supervised & Vim-Base~\cite{vim} & Mamba~\cite{mamba} & IN1k & class & 448$\times$448 & 73.3 GB & 56 & 83.5 \\
    supervised & Adventurer-Base~\cite{adventurer} & Mamba-2~\cite{mamba2} & IN1k & class & 448$\times$448 & 45.2 GB & 199 & 84.2 \\\rowcolor{orange!10}
    ViT-Linearizer & Adventurer-Base~\cite{adventurer} & Mamba-2~\cite{mamba2} & IN1k & CLIP & 448$\times$448 & 45.2 GB & 199 & \bf 85.0 \\
\end{tabular}
\caption{\textbf{ImageNet-1k classification results.} We fine-tune the listed models and show their efficiency and accuracy comparisons. Following~\cite{adventurer}, we measure the memory usage at a fixed batch size of 128 per GPU during training. Throughput (images/second) is evaluated on a single RTX4090 GPU. Models trained with data beyond ImageNet are \demph{de-emphasized}. Our results are highlighted in \textcolor{orange}{orange}.}
\label{tab:cls}
\end{table*}

\paragraph{ImageNet classification.} Our ViT-Linearizer approach demonstrates superior fine-tuning performance on ImageNet-1k, particularly when compared to existing Mamba-based models such as Vim and supervised Adventurer. As summarized in Table~\ref{tab:cls}, when using a 224$\times$224 input size, our distilled Adventurer-Base outperforms previous supervised models such as Vim-B~\cite{vim}, DeiT-B~\cite{deit}, and even the state-of-the-art ViT-B from DeiT-III, highlighting the great effectiveness of our cross-architecture knowledge transfer strategy.

One key factor contributing to this success is our choice of a strong teacher model: the ViT-Base/16 from CLIP, which is pretrained on a significantly larger dataset~\cite{clip} (400M image-text pairs) compared to ImageNet-1k. Notably, despite its reduced computational complexity, our distilled model exhibits minimal performance degradation relative to directly fine-tuning the teacher model. For example, in experiments with a 448$\times$448 input size, our model achieves a 2.1$\times$ inference speedup while sacrificing only 0.3\% accuracy, which we consider as a clear indication of the effectiveness of our approach in balancing efficiency and accuracy. Importantly, the efficiency gains of ViT-Linearizer scale with sequence length, making it increasingly advantageous in long-context scenarios. For example, in our ablation study, refining the model granularity by using a smaller patch size (\eg, 8$\times$8) during fine-tuning enhances the acceleration factor from 2.1$\times$ to 4.6$\times$. It is noteworthy that this speed improvement has no upper bound, since the advantage of our approach becomes more pronounced as sequence length increases. This suggests that ViT-Linearizer holds significant potential for efficiently handling high-resolution and long-context vision tasks.

\paragraph{Semantic Segmentation.} We also evaluate the effectiveness of ViT-Linearizer for the semantic segmentation tasks, where input sizes are typically larger than those used for classification (\eg, 512$\times$512 for ADE20k~\cite{ade20k} and 512$\times$1024 for Cityscapes~\cite{cityscapes}). As shown in Table~\ref{tab:ade}, our Adventurer-Base, which is distilled from CLIP's ViT-B/16 through ViT-Linearizer, achieves superior performance compared to similarly sized baseline models on the ADE20k benchmark. Notably, the distilled model not only improves inference speed by 2.74$\times$ over its ViT teacher but also achieves a slightly higher test mIoU (51.3\% \emph{vs.} 51.0\%), further validating the previous findings that Mamba-based recurrent vision models excel in long-sequence dense prediction tasks \textbf{\textit{for both efficiency and accuracy}}~\cite{vim,mambaout,vmamba}.

We also evaluate semantic segmentation performance on the Cityscapes dataset~\cite{cityscapes}, which features higher input resolution and consequently longer input sequences. In this benchmark, images are typically cropped to 512×1024 pixels, resulting in a 2K-length visual sequence under the standard 16×16-pixel patchification process. At this sequence length, the computational cost of ViTs becomes increasingly dominated by self-attention, making them highly sensitive to the sequence length, suggesting that the efficiency gains obtained by our ViT-Linearization approach become even more pronounced---we achieve a remarkable speedup of 4.21$\times$ inference throughput, while the predictive performance \textbf{\textit{does not experience any degradations.}}

Moreover, we would like to highlight that a 2K-length sequence remains an initial step in exploring ViT linearization. In this work, we demonstrate the practical benefits of our method on commonly used medium-scale visual understanding tasks, within manageable computational resources. However, we believe its potential extends far beyond this: recent studies have pushed for super-fine-grained patchification, increasing the token count per image to 50K or even more, with consistent performance gains from longer sequences~\cite{patchscale}. We leave the exploration of larger-scale ViT-Linearizer models for future work and anticipate that they will play more critical roles in long-sequence vision tasks.

\begin{table}[t]
    \centering
    \tablestyle{5pt}{1.1}
    \begin{tabular}{l|rrr}
    Backbone & Parameters & Throughput ($\uparrow$) & mIoU (\%) \\\shline
    \demph{CLIP ViT-B/16~\cite{clip}} & \demph{119M} & \demph{1.00$\times$} & \demph{51.0} \\
    DeiT-Base~\cite{deit} & 119M & 1.00$\times$ & 45.5 \\
    DeiT-III-Base~\cite{deit3} & 120M & 0.96$\times$ & 49.3 \\
    MAE ViT-Base~\cite{mae} & 119M & 1.00$\times$ & 48.5 \\\hline
    Vim-Base~\cite{vim} & 132M & 0.68$\times$ & 45.5 \\
    Adventurer-Base~\cite{adventurer} & 115M & 2.74$\times$ & 47.8 \\\rowcolor{orange!10}
    Adventurer-Base (ours) & 115M & \bf 2.74$\times$ & \bf 51.3 \\
    
    \end{tabular}
    \caption{\textbf{ADE20k semantic segmentation results.} We employ an UperNet~\cite{upernet} decoder head for all models, with the input size fixed to 512$\times$512. To eliminate the influence of decoders on inference time, the throughput is calculated based solely on the encoding time consumption. Our model is distilled from CLIP ViT-B/16.}
    \label{tab:ade}
\end{table}

\begin{table}[t]
    \centering
    \tablestyle{5pt}{1.1}
    \begin{tabular}{l|rrr}
    Backbone & Parameters & Throughput ($\uparrow$) & mIoU (\%) \\\shline
    \demph{CLIP ViT-B/16~\cite{clip}} & \demph{122M} & \demph{1.00$\times$} & \demph{81.8} \\
    DeiT-Base~\cite{deit} & 122M & 1.00$\times$ & 79.6 \\
    DeiT-III-Base~\cite{deit3} & 123M & 0.97$\times$ & 80.8 \\
    MAE ViT-Base~\cite{mae} & 122M & 1.00$\times$ & 80.9 \\\hline
    Vim-Base~\cite{vim} & 135M & 1.12$\times$ & 79.2 \\
    Adventurer-Base~\cite{adventurer} & 118M & 4.21$\times$ & 80.2 \\\rowcolor{orange!10}
    Adventurer-Base (ours) & 118M & \bf 4.21$\times$ & \bf 82.0 \\
    
    \end{tabular}
    \caption{\textbf{Cityscapes semantic segmentation results.} Similar to Table~\ref{tab:ade}, we employ an UperNet~\cite{upernet} decoder head and measure throughput by only encoding time. The input size is 512$\times$1024 resolution. Our model is distilled from CLIP ViT-Base/16.}
    \label{tab:city}
\end{table}

\begin{figure*}[t]
    \centering
    \includegraphics[width=0.95\textwidth]{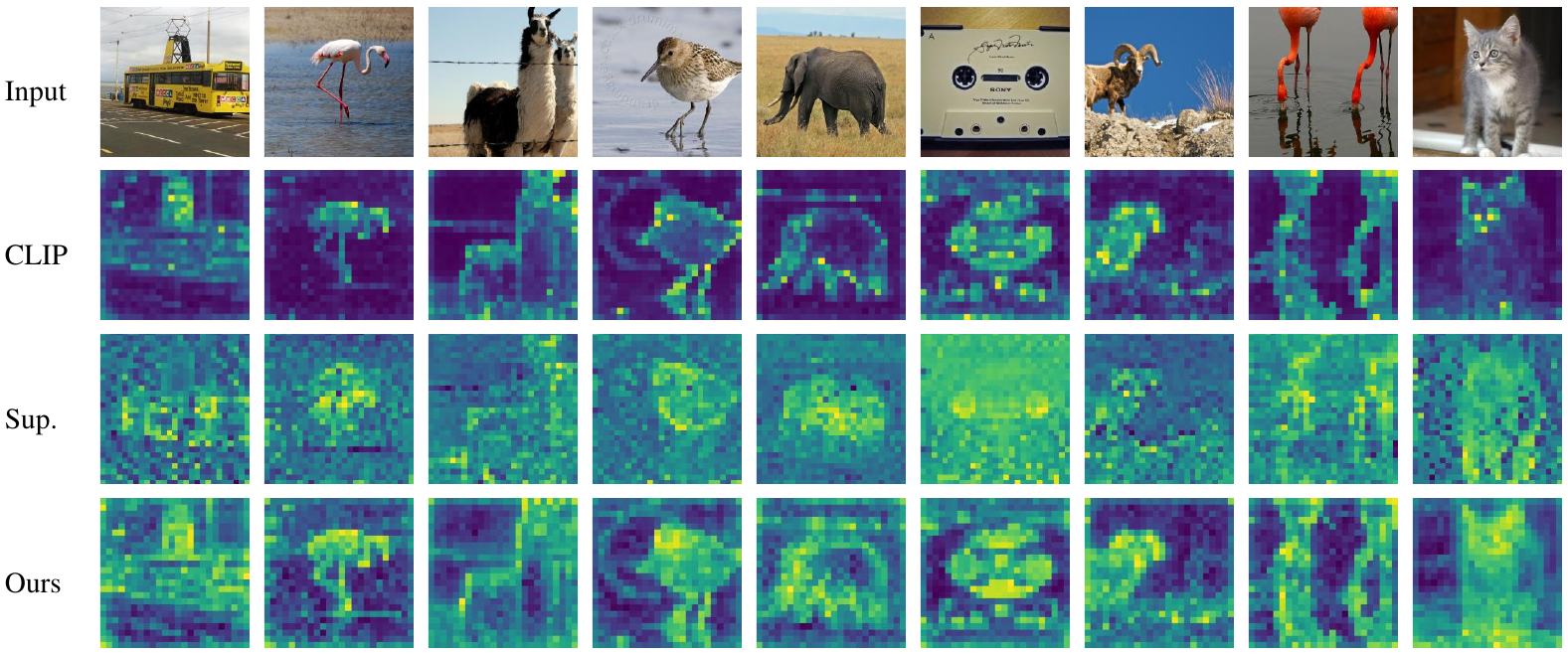}
    \caption{\textbf{Qualitative comparison of activation maps.} The teacher model (CLIP ViT-B/16) consistently produces high-contrast activations with distinctly highlighted salient regions. The supervised Adventurer baseline (denoted ``Sup.'') exhibits noisy activations. Our distilled Adventurer model shows significant improvements, with feature patterns closely aligning to those of its ViT teacher.}
    \label{fig:attn}
\end{figure*}

\subsection{Qualitative Analysis}

In Figure~\ref{fig:attn}, we present a qualitative comparison of the activation maps produced by the teacher ViT (CLIP ViT-B/16), a purely supervised Adventurer model, and our Adventurer model distilled via ViT-Linearizer. Consistent with many self-attention-based models, the CLIP ViT model exhibits high-contrast activation patterns across diverse inputs, where informative foreground tokens receive strong activation scores and object boundaries are well delineated. Similar to the observations and analyses in prior works~\cite{dino,dinov2,register,mambar}, we consider such distinct feature maps as a direct reflection of how self-attention leverages its quadratic computational complexity to obtain clear local representations---in self-attention, each token is required to compute correlations with all other positions, which notably facilitates filtering out tokens that are minimally relevant to the current context.

By contrast, the recurrent models naturally face limitations in filtering out non-informative tokens due to their reduced computational complexity, which is typically reflected by lower-contrast and noisier activation maps compared with ViTs. Yet interestingly, by compelling the recurrent model to directly learn the ViT’s activations (\ie, the activation matching mechanism), our cross-architecture distillation framework produces not only substantial performance gains in quantitative metrics, but also significantly clearer activation maps that are more closely aligned with those of the ViT teachers. Notably, these results provide evidence that our approach indeed enables recurrent models like Adventurer to inherit the representational knowledge acquired by ViTs at quadratic cost, and effectively replicate the teacher’s sharper activation patterns despite the inherent constraints of a linear-time formulation.

\subsection{Ablation Study}
\label{sec:abl}

\begin{table*}[t]
    \centering
    \tablestyle{5pt}{1.1}
    \begin{tabular}{lccccc|lccc}
    \multicolumn{5}{c}{Teacher model} & & \multicolumn{4}{c}{Student model (ours)} \\

    model & pt. data & supervision & img. size & throughput & acc.(\%) & model & img. size & throughput & acc.(\%) \\\shline
    
    \multirow{2}{*}{MAE ViT-B~\cite{mae}} & \multirow{2}{*}{IN1k} & \multirow{2}{*}{pixel} & 224$\times$224 & 612 & 83.6 & \multirow{2}{*}{Adventurer-B~\cite{adventurer}} & 224$\times$224 & 736 \scriptsize \gtext{(1.20$\times$)} & 83.4 \\
    &&& 448$\times$448 & 95 & 84.9 & & 448$\times$448 & 199 \scriptsize \gtext{(2.09$\times$)} & 84.5 \\\hline

    \multirow{2}{*}{DeiT-III ViT-B~\cite{deit3}} & \multirow{2}{*}{IN1k} & \multirow{2}{*}{class} & 224$\times$224 & 608 & 83.8 & \multirow{2}{*}{Adventurer-B~\cite{adventurer}} & 224$\times$224 & 736 \scriptsize \gtext{(1.21$\times$)} & 83.6 \\
    &&& 448$\times$448 & 94 & 85.0 & & 448$\times$448 & 199 \scriptsize \gtext{(2.12$\times$)} & 84.6 \\\hline

    \multirow{2}{*}{CLIP ViT-B~\cite{clip}} & \multirow{2}{*}{WIT-400M} & \multirow{2}{*}{text} & 224$\times$224 & 613 & 84.7 & \multirow{2}{*}{Adventurer-B~\cite{adventurer}} & 224$\times$224 & 736 \scriptsize \gtext{(1.20$\times$)} & 84.3 \\
    &&& \cellcolor{orange!10} 448$\times$448 & \cellcolor{orange!10} 95 & \cellcolor{orange!10} \bf 85.3 & & \cellcolor{orange!10} 448$\times$448 & \cellcolor{orange!10} 199 \scriptsize \gtext{(2.09$\times$)} & \cellcolor{orange!10} \bf 85.0 \\
    \end{tabular}
    \caption{\textbf{ViT-Linearizer with various teacher models.} We evaluate the effectiveness of our ViT-Linearizer by distilling from teacher models learned by different pretraining paradigms. We keep the teacher (ViT-Base) and student (Adventurer-Base) models in a similar size of parameters. The measurement of throughput follows the same protocol of Table~\ref{tab:cls}.}
    \label{tab:abl_tea}
\end{table*}

\paragraph{Distill from different teachers.} As shown in Table~\ref{tab:abl_tea}, we first validate the generalizability of our ViT-Linearizer approach by distilling knowledge from three state-of-the-art ViT teachers, each representing a different pretraining paradigm: the fully supervised DeiT-III~\cite{deit3}, unsupervised MAE~\cite{mae}, and the weakly supervised CLIP with text-based targets. These models collectively span a broad range of current pretraining strategies. Our results show that ViT-Linearizer consistently performs well across all the three teacher models, with the student’s downstream finetuning accuracy closely reflecting the data scale of its respective teacher. In particular, the best performance is observed when distilling from CLIP, which benefits from a large-scale pretraining corpus. Notably, in every case, ViT-Linearizer achieves a substantial speedup (\eg over 2$\times$ for the 448-sized inputs) with less than a 0.5\% drop in accuracy, indicating that the approach delivers significant acceleration with only marginal loss in predictive performance, regardless of the origin of teacher models. These findings highlight the versatility of ViT-Linearizer, suggesting it as a widely applicable solution for efficiently accelerating a broad range of pretrained ViT models without compromising performance.

\begin{table}[t]
    \centering
    \tablestyle{5pt}{1.1}
    \begin{tabular}{lr|lrr}
    Teacher & Params & Student & Params & Acc. (\%) \\\shline
    \multirow{3}{*}{CLIP ViT-B/16} & \multirow{3}{*}{86M} & Adventurer-S/16 & 44M & 83.1 \\
    & & Adventurer-B/16 & 99M & 84.3 \\
    & & \cellcolor{orange!10} Adventurer-L/16 & \cellcolor{orange!10} 346M & \cellcolor{orange!10} \bf 85.0 \\\hline
    \multirow{3}{*}{CLIP ViT-L/14} & \multirow{3}{*}{307M} & Adventurer-S/14 & 44M & 83.4 \\
    & & Adventurer-B/14 & 99M & 84.5 \\
    & & Adventurer-L/14 & 346M & \bf 85.2 \\
    \end{tabular}
    \caption{\textbf{Distillation across different model sizes.} Models are evaluated on ImageNet-1k. We observe an ``inverse'' distillation phenomenon that a Large-szied student model can also benefit from a Base-sized teacher (highlighted in \textcolor{orange}{orange}).}
    \label{tab:abl_stu}
\end{table}

\paragraph{Student model sizes.} We next investigate the effect of student model size on distillation performance. As shown in Table~\ref{tab:abl_stu}, we compare Adventurer variants of different parameter scales distilled from either CLIP ViT-B/16 (Base teacher) or CLIP ViT-L/14 (Large teacher). Interestingly, we find that the sizes of the teacher and student do not need to be strictly aligned. In addition to the typical scenario of transferring knowledge from a larger teacher to a smaller student, both similarly sized and even ``inversely sized'' (where the student exceeds the teacher in parameter count) configurations also yield strong results. Notably, we successfully train an Adventurer-L/14 model to 85.2\% top-1 accuracy on ImageNet---surpassing the previous supervised Adventurer-Large record of 83.4\%---thus establishing a new state-of-the-art for this family of recurrent models. We interpret these results to indicate that ViT-Linearizer not only helps to reduce inference costs for ViTs, but also endows recurrent vision architectures like Adventurer with ``attentive'' knowledge and masked image modeling capabilities, both widely recognized as effective means of enhancing visual understanding.

\begin{table}[t]
    \centering
    \tablestyle{5pt}{1.1}
    \begin{tabular}{l|cccc}
    Mode & Mask pred. & Act. match & IN1k acc. & ADE mIoU \\\shline
    \demph{supervised} & \demph{\xmark} & \demph{\xmark} & \demph{82.6} & \demph{47.8} \\
    no act. match & \cmark & \xmark & 83.6 & 49.7\\
    no mask pred. & \xmark & \cmark & 83.8 & 50.1 \\\rowcolor{orange!10}
    default & \cmark & \cmark & \bf 84.3 & \bf 51.3 \\
    \end{tabular}
    \caption*{(a) Components of masked prediction and activation matching.}

    \vspace{+0.5cm}

    \begin{tabular}{l|cc}
    Mode & Block-wise masking~\cite{beit} & Token-wise masking~\cite{mae} \\\shline
    IN1k acc. & 84.0 & \cellcolor{orange!10} 84.3 \\
    ADE mIoU & 50.6 & \cellcolor{orange!10} 51.3\\
    \end{tabular}
    \caption*{(b) Masking strategies. Our default setup is highlighted in \textcolor{orange}{orange}.}

    \vspace{+0.5cm}

    \begin{tabular}{l|ccc}
    Mode & Class token only & Visible tokens only & All tokens \\\shline
    IN1k acc. & 83.7 & \cellcolor{orange!10} 84.3 & 83.4 \\
    ADE mIoU & 50.0 & \cellcolor{orange!10} 51.3 & 49.0 \\
    \end{tabular}
    \caption*{(c) Activation matching details. Our default setup is highlighted.}
    
    \caption{\textbf{Ablation of key components.} We compare different modes by their ImageNet accuracy (\%) and ADE20k mIoU (\%).}
    \label{tab:component}
\end{table}

\paragraph{Designing details.} As shown in Table~\ref{tab:component}, we ablate the key components of our approach to examine their individual and combined effects on both image-level (ImageNet) and pixel-level (ADE20k) tasks. We first observe that applying either activation matching or masked prediction alone brings noticeable gains, yet neither approach in isolation achieves our best results across both benchmarks. Regarding masked prediction, we compare two common strategies---block-wise masking as in BEiT~\cite{beit} versus token-wise random masking as in MAE~\cite{mae}---and find that the latter offers greater flexibility and consistently higher finetuning performance. We further investigate different activation matching setups to ensure proper synergy with masked prediction. As discussed in Section~\ref{sec:mask}, our default practice matches only visible tokens, preventing potential information leakage from the final-layer masked tokens. Empirically, matching all tokens notably degrades performance, confirming our initial assumption. Another simplified variant is to match only the class token, which reduces the complexity of $\mathcal{L}_\text{act}$ to $\mathcal{O}(L)$ where $L$ denotes sequence length. However, as shown in Table~\ref{tab:component} (c), this weaker constraint hinders knowledge transfer and leads to lower accuracy. Taken together, these findings affirm the importance of a stronger constraint on all visible tokens to fully exploit the ``quadratic knowledge'' learned by the teacher model.

%% file: sec/5_conclusion.tex
\section{Conclusion}

In this paper, we introduce ViT-Linearizer, a cross-architecture distillation approach that transfers high-capacity ViT representations to linear-time, RNN-based vision models such as Adventurer. By aligning token activations and integrating masked prediction, the Mamba-based Adventurer models effectively inherit the quadratic knowledge from self-attention while reducing inference cost. Extensive experiments on both image-level and pixel-level tasks verify that our method achieves competitive or superior performance relative to baseline recurrent architectures at a fraction of the computational overhead. Moreover, ViT-Linearizer generalizes across diverse teacher models (e.g., DeiT-III, MAE, CLIP) and different student capacities, underscoring its versatility. We envision this paradigm as a bridge between large-scale Transformers and more efficient architectures in real-world, resource-constrained settings. As the demand for high-resolution image processing continues to surge, the advantages of linear-time models will grow increasingly critical, further amplifying the practical impact of our approach.

%% file: sec/acknowledge.tex
\section{Acknowledgements}

The authors GW and RC were supported by an ONR MURI grant N00014-20-1-2787.

%% file: sec/X_Appendix.tex
\newpage
\onecolumn

\section*{Appendix}

\subsection*{A. Technical Details}

\paragraph{Model configurations.} In all our experiments, we use ViT as the teacher model and Adventurer as the student model—both featuring a plain (non-hierarchical) design that maintains consistent spatial resolutions across layers. Their detailed configurations are summarized in Table~\ref{tab:config}.

\begin{table}[h!]
    \tablestyle{5pt}{1.1}
    \begin{tabular}{l|cccc}
        Model & Embedding dimension & MLP dimension & Blocks & Parameters \\\shline
        ViT-Base, Patch size 16$\times$16 & 768 & 3,072 & 12 & 86M \\
        ViT-Large, Patch size 14$\times$14 & 1,024 & 4,096 & 24 & 307M \\\hline
        Adventurer-Small, Patch size 16$\times$16 & 512 & 1,280 & 12 & 44M \\
        Adventurer-Base, Patch size 16$\times$16 & 768 & 1,920 & 12 & 99M \\
        Adventurer-Large, Patch size 14$\times$14 & 1,024 & 2,560 & 24 & 346M \\
    \end{tabular}

    \caption{Detailed configuration of the models used in this paper.}
    \label{tab:config}
\end{table}

\paragraph{Training recipes.} In our distillation stage, we did not perform extensive hyperparameter tuning. Instead, we mainly followed the settings adopted in prior ViT-based masked distillation studies~\cite{maskdistill}, but applied stronger data augmentation and higher drop path rates, which previous findings~\cite{mambar,adventurer} suggest are better suited for Mamba-style models. Detailed hyper=parameters can be found in Table~\ref{tab:recipe_pt} and ~\ref{tab:recipe_ft}. For semantic segmentation fine-tuning, we simply follow the recipe in~\cite{maskdistill}.

\begin{table}[h]
    \centering
    \tablestyle{10pt}{1.1}
    \begin{tabular}{l|cc}
    Config & Small/Base & Large \\\shline
    optimizer & \multicolumn{2}{c}{AdamW} \\
    peak learning rate & \multicolumn{2}{c}{1.5e-3} \\
    minimum learning rate & \multicolumn{2}{c}{1e-5} \\
    weight decay & \multicolumn{2}{c}{0.05} \\
    epochs & \multicolumn{2}{c}{300} \\
    optimizer betas & \multicolumn{2}{c}{0.9, 0.999} \\
    batch size & \multicolumn{2}{c}{2048} \\
    warmup epochs & 10 & 20 \\
    stochastic depth (drop path) & 0.1 & 0.2 \\
    layer-wise lr decay & \multicolumn{2}{c}{\xmark} \\
    label smoothing & \multicolumn{2}{c}{\xmark} \\
    random erasing  & \multicolumn{2}{c}{\xmark} \\
    Rand Augmentation & \multicolumn{2}{c}{\xmark} \\
    repeated augmentation & \multicolumn{2}{c}{\cmark} \\
    ThreeAugmentation & \multicolumn{2}{c}{\cmark} \\
    \end{tabular}
    \caption{Configurations of the distillation stage.}
    \label{tab:recipe_pt}
\end{table}

\begin{table}[h]
    \centering
    \tablestyle{10pt}{1.1}
    \begin{tabular}{l|cc}
    Config & Small/Base & Large \\\shline
    optimizer & \multicolumn{2}{c}{AdamW} \\
    peak learning rate & \multicolumn{2}{c}{5e-4} \\
    minimum learning rate & \multicolumn{2}{c}{1e-6} \\
    weight decay & \multicolumn{2}{c}{0.05} \\
    epochs & 100 & 50 \\
    optimizer betas & \multicolumn{2}{c}{0.9, 0.999} \\
    batch size & \multicolumn{2}{c}{1024} \\
    warmup epochs & 20 & 5 \\
    stochastic depth (drop path) & 0.4 & 0.6 \\
    layer-wise lr decay & 0.65 & 0.8 \\
    label smoothing & \multicolumn{2}{c}{\cmark} \\
    random erasing  & \multicolumn{2}{c}{\xmark} \\
    Rand Augmentation & \multicolumn{2}{c}{rand-m9-mstd0.5-inc1} \\
    \end{tabular}
    \caption{Configurations of the fine-tuning stage.}
    \label{tab:recipe_ft}
\end{table}

%% file: main.bib
@String(IJCV  = {IJCV})

@String(CVPR  = {CVPR})

@String(ICCV  = {ICCV})

@String(ECCV  = {ECCV})

@String(NIPS  = {NeurIPS})

@String(ICLR  = {ICLR})

@String(AAAI = {AAAI})

@String(icml = "{ICML}")

@String(nips = {NeurIPS})

@article{adventurer,
  title={Causal Image Modeling for Efficient Visual Understanding},
  author={Wang, Feng and Yang, Timing and Yu, Yaodong and Ren, Sucheng and Wei, Guoyizhe and Wang, Angtian and Shao, Wei and Zhou, Yuyin and Yuille, Alan and Xie, Cihang},
  journal={arXiv preprint arXiv:2410.07599},
  year={2024}
}

@inproceedings{dit,
  title={Scalable diffusion models with transformers},
  author={Peebles, William and Xie, Saining},
  booktitle={ICCV},
  year={2023}
}

@inproceedings{sclip,
  title={Sclip: Rethinking self-attention for dense vision-language inference},
  author={Wang, Feng and Mei, Jieru and Yuille, Alan},
  booktitle={ECCV},
  year={2024}
}

@article{arm,
  title={Autoregressive Pretraining with Mamba in Vision},
  author={Ren, Sucheng and Li, Xianhang and Tu, Haoqin and Wang, Feng and Shu, Fangxun and Zhang, Lei and Mei, Jieru and Yang, Linjie and Wang, Peng and Wang, Heng and others},
  journal={arXiv preprint arXiv:2406.07537},
  year={2024}
}

@article{mambar,
  title={Mamba-r: Vision mamba also needs registers},
  author={Wang, Feng and Wang, Jiahao and Ren, Sucheng and Wei, Guoyizhe and Mei, Jieru and Shao, Wei and Zhou, Yuyin and Yuille, Alan and Xie, Cihang},
  journal={arXiv preprint arXiv:2405.14858},
  year={2024}
}

@inproceedings{mamba2,
  title={Transformers are SSMs: Generalized Models and Efficient Algorithms Through Structured State Space Duality},
  author={Dao, Tri and Gu, Albert},
  booktitle={ICML},
  year={2024}
}

@article{llama,
  title={Llama: Open and efficient foundation language models},
  author={Touvron, Hugo and Lavril, Thibaut and Izacard, Gautier and Martinet, Xavier and Lachaux, Marie-Anne and Lacroix, Timoth{\'e}e and Rozi{\`e}re, Baptiste and Goyal, Naman and Hambro, Eric and Azhar, Faisal and others},
  journal={arXiv preprint arXiv:2302.13971},
  year={2023}
}

@article{defo,
  title={Learning to decompose visual features with latent textual prompts},
  author={Wang, Feng and Li, Manling and Lin, Xudong and Lv, Hairong and Schwing, Alexander G and Ji, Heng},
  journal={arXiv preprint arXiv:2210.04287},
  year={2022}
}

@inproceedings{clip,
  title={Learning transferable visual models from natural language supervision},
  author={Radford, Alec and Kim, Jong Wook and Hallacy, Chris and Ramesh, Aditya and Goh, Gabriel and Agarwal, Sandhini and Sastry, Girish and Askell, Amanda and Mishkin, Pamela and Clark, Jack and others},
  booktitle={ICML},
  year={2021}
}

@inproceedings{maskclip,
  title={Extract free dense labels from clip},
  author={Zhou, Chong and Loy, Chen Change and Dai, Bo},
  booktitle=eccv,
  year={2022}
}

@inproceedings{mae,
  title={Masked autoencoders are scalable vision learners},
  author={He, Kaiming and Chen, Xinlei and Xie, Saining and Li, Yanghao and Doll{\'a}r, Piotr and Girshick, Ross},
  booktitle={CVPR},
  year={2022}
}

@inproceedings{beit,
title={{BEiT}: {BERT} Pre-Training of Image Transformers}, 
  author={Bao, Hangbo and Dong, Li and Wei, Furu},
  booktitle={ICLR},
  year={2022}
}

@inproceedings{maskfeat,
  title={Masked feature prediction for self-supervised visual pre-training},
  author={Wei, Chen and Fan, Haoqi and Xie, Saining and Wu, Chao-Yuan and Yuille, Alan and Feichtenhofer, Christoph},
  booktitle=cvpr,
  year={2022}
}

@inproceedings{dino,
  title={Emerging Properties in Self-Supervised Vision Transformers},
  author={Caron, Mathilde and Touvron, Hugo and Misra, Ishan and J\'egou, Herv\'e  and Mairal, Julien and Bojanowski, Piotr and Joulin, Armand},
  booktitle={ICCV},
  year={2021}
}

@inproceedings{vit,
  title={An Image is Worth 16x16 Words: Transformers for Image Recognition at Scale},
  author={Dosovitskiy, Alexey and Beyer, Lucas and Kolesnikov, Alexander and Weissenborn, Dirk and Zhai, Xiaohua and Unterthiner, Thomas and  Dehghani, Mostafa and Minderer, Matthias and Heigold, Georg and Gelly, Sylvain and Uszkoreit, Jakob and Houlsby, Neil},
  booktitle={ICLR},
  year={2021}
}

@inproceedings{imagenet,
  title={{ImageNet}: A large-scale hierarchical image database},
  author={Deng, Jia and Dong, Wei and Socher, Richard and Li, Li-Jia and Li, Kai and Fei-Fei, Li},
  booktitle={CVPR},
  year={2009}
}

@article{voc12,
title={The pascal visual object classes challenge: A retrospective},
author={Everingham, Mark and Eslami, SM Ali and Van Gool, Luc and Williams, Christopher KI and Winn, John and Zisserman, Andrew},
journal={IJCV},
year={2015}
}

@inproceedings{pascalcontext,
  title={The role of context for object detection and semantic segmentation in the wild},
  author={Mottaghi, Roozbeh and Chen, Xianjie and Liu, Xiaobai and Cho, Nam-Gyu and Lee, Seong-Whan and Fidler, Sanja and Urtasun, Raquel and Yuille, Alan},
  booktitle=cvpr,
  year={2014}
}

@inproceedings{coco,
  title={Coco-stuff: Thing and stuff classes in context},
  author={Caesar, Holger and Uijlings, Jasper and Ferrari, Vittorio},
  booktitle=cvpr,
  year={2018}
}

@inproceedings{cityscapes,
  title={The cityscapes dataset for semantic urban scene understanding},
  author={Cordts, Marius and Omran, Mohamed and Ramos, Sebastian and Rehfeld, Timo and Enzweiler, Markus and Benenson, Rodrigo and Franke, Uwe and Roth, Stefan and Schiele, Bernt},
  booktitle=cvpr,
  year={2016}
}

@article{ade20k,
  title={Semantic understanding of scenes through the ade20k dataset},
  author={Zhou, Bolei and Zhao, Hang and Puig, Xavier and Xiao, Tete and Fidler, Sanja and Barriuso, Adela and Torralba, Antonio},
  journal=ijcv,
  year={2019}
}

@inproceedings{stablediff,
  title={High-resolution image synthesis with latent diffusion models},
  author={Rombach, Robin and Blattmann, Andreas and Lorenz, Dominik and Esser, Patrick and Ommer, Bj{\"o}rn},
  booktitle=cvpr,
  year={2022}
}

@article{llava,
  title={Visual instruction tuning},
  author={Liu, Haotian and Li, Chunyuan and Wu, Qingyang and Lee, Yong Jae},
  journal={arXiv preprint arXiv:2304.08485},
  year={2023}
}

@inproceedings{deit,
  title={Training data-efficient image transformers \& distillation through attention},
  author={Touvron, Hugo and Cord, Matthieu and Douze, Matthijs and Massa, Francisco and Sablayrolles, Alexandre and J{\'e}gou, Herv{\'e}},
  booktitle=icml,
  year={2021}
}

@inproceedings{deit3,
  title={Deit iii: Revenge of the vit},
  author={Touvron, Hugo and Cord, Matthieu and J{\'e}gou, Herv{\'e}},
  booktitle=eccv,
  year={2022}
}

@article{rwkv,
  title={Rwkv: Reinventing rnns for the transformer era},
  author={Peng, Bo and Alcaide, Eric and Anthony, Quentin and Albalak, Alon and Arcadinho, Samuel and Cao, Huanqi and Cheng, Xin and Chung, Michael and Grella, Matteo and GV, Kranthi Kiran and others},
  journal={arXiv preprint arXiv:2305.13048},
  year={2023}
}

@article{mamba,
  title={Mamba: Linear-time sequence modeling with selective state spaces},
  author={Gu, Albert and Dao, Tri},
  journal={arXiv preprint arXiv:2312.00752},
  year={2023}
}

@inproceedings{vim,
  title={Vision mamba: Efficient visual representation learning with bidirectional state space model},
  author={Zhu, Lianghui and Liao, Bencheng and Zhang, Qian and Wang, Xinlong and Liu, Wenyu and Wang, Xinggang},
  booktitle={ICML},
  year={2024}
}

@article{vmamba,
  title={Vmamba: Visual state space model},
  author={Liu, Yue and Tian, Yunjie and Zhao, Yuzhong and Yu, Hongtian and Xie, Lingxi and Wang, Yaowei and Ye, Qixiang and Liu, Yunfan},
  journal={arXiv preprint arXiv:2401.10166},
  year={2024}
}

@article{videomamba,
  title={Videomamba: State space model for efficient video understanding},
  author={Li, Kunchang and Li, Xinhao and Wang, Yi and He, Yinan and Wang, Yali and Wang, Limin and Qiao, Yu},
  journal={arXiv preprint arXiv:2403.06977},
  year={2024}
}

@inproceedings{register,
  title={Vision transformers need registers},
  author={Darcet, Timoth{\'e}e and Oquab, Maxime and Mairal, Julien and Bojanowski, Piotr},
  booktitle={ICLR},
  year={2024}
}

@inproceedings{adamw,
  title={Decoupled weight decay regularization},
  author={Loshchilov, Ilya and Hutter, Frank},
  booktitle=iclr,
  year={2019}
}

@inproceedings{upernet,
  title={Unified perceptual parsing for scene understanding},
  author={Xiao, Tete and Liu, Yingcheng and Zhou, Bolei and Jiang, Yuning and Sun, Jian},
  booktitle=eccv,
  year={2018}
}

@article{plainmamba,
  title={Plainmamba: Improving non-hierarchical mamba in visual recognition},
  author={Yang, Chenhongyi and Chen, Zehui and Espinosa, Miguel and Ericsson, Linus and Wang, Zhenyu and Liu, Jiaming and Crowley, Elliot J},
  journal={arXiv preprint arXiv:2403.17695},
  year={2024}
}

@article{mambavision,
  title={MambaVision: A Hybrid Mamba-Transformer Vision Backbone},
  author={Hatamizadeh, Ali and Kautz, Jan},
  journal={arXiv preprint arXiv:2407.08083},
  year={2024}
}

@misc{dinov2,
  title={DINOv2: Learning Robust Visual Features without Supervision},
  author={Oquab, Maxime and Darcet, Timothée and Moutakanni, Theo and Vo, Huy V. and Szafraniec, Marc and Khalidov, Vasil and Fernandez, Pierre and Haziza, Daniel and Massa, Francisco and El-Nouby, Alaaeldin and Howes, Russell and Huang, Po-Yao and Xu, Hu and Sharma, Vasu and Li, Shang-Wen and Galuba, Wojciech and Rabbat, Mike and Assran, Mido and Ballas, Nicolas and Synnaeve, Gabriel and Misra, Ishan and Jegou, Herve and Mairal, Julien and Labatut, Patrick and Joulin, Armand and Bojanowski, Piotr},
  journal={arXiv:2304.07193},
  year={2023}
}

@article{gpt4,
  title={Gpt-4 technical report},
  author={Achiam, Josh and Adler, Steven and Agarwal, Sandhini and Ahmad, Lama and Akkaya, Ilge and Aleman, Florencia Leoni and Almeida, Diogo and Altenschmidt, Janko and Altman, Sam and Anadkat, Shyamal and others},
  journal={arXiv preprint arXiv:2303.08774},
  year={2023}
}

@article{retnet,
  title={Retentive network: A successor to transformer for large language models},
  author={Sun, Yutao and Dong, Li and Huang, Shaohan and Ma, Shuming and Xia, Yuqing and Xue, Jilong and Wang, Jianyong and Wei, Furu},
  journal={arXiv preprint arXiv:2307.08621},
  year={2023}
}

@article{xlstm,
  title={xLSTM: Extended Long Short-Term Memory},
  author={Beck, Maximilian and P{\"o}ppel, Korbinian and Spanring, Markus and Auer, Andreas and Prudnikova, Oleksandra and Kopp, Michael and Klambauer, G{\"u}nter and Brandstetter, Johannes and Hochreiter, Sepp},
  journal={arXiv preprint arXiv:2405.04517},
  year={2024}
}

@article{visionrwkv,
  title={Vision-rwkv: Efficient and scalable visual perception with rwkv-like architectures},
  author={Duan, Yuchen and Wang, Weiyun and Chen, Zhe and Zhu, Xizhou and Lu, Lewei and Lu, Tong and Qiao, Yu and Li, Hongsheng and Dai, Jifeng and Wang, Wenhai},
  journal={arXiv preprint arXiv:2403.02308},
  year={2024}
}

@article{visionlstm,
  title={Vision-LSTM: xLSTM as Generic Vision Backbone},
  author={Alkin, Benedikt and Beck, Maximilian and P{\"o}ppel, Korbinian and Hochreiter, Sepp and Brandstetter, Johannes},
  journal={arXiv preprint arXiv:2406.04303},
  year={2024}
}

@article{distillquadra,
  title={Transformers to ssms: Distilling quadratic knowledge to subquadratic models},
  author={Bick, Aviv and Li, Kevin and Xing, Eric and Kolter, J Zico and Gu, Albert},
  journal=nips,
  year={2025}
}

@article{maskdistill,
  title={A Unified View of Masked Image Modeling},
  author={Peng, Zhiliang and Dong, Li and Bao, Hangbo and Ye, Qixiang and Wei, Furu},
  journal={arXiv preprint arXiv:2406.04303},
  year={2022}
}

@inproceedings{vqav2,
  title={Making the v in vqa matter: Elevating the role of image understanding in visual question answering},
  author={Goyal, Yash and Khot, Tejas and Summers-Stay, Douglas and Batra, Dhruv and Parikh, Devi},
  booktitle=cvpr,
  year={2017}
}

@article{ibot,
  title={ibot: Image bert pre-training with online tokenizer},
  author={Zhou, Jinghao and Wei, Chen and Wang, Huiyu and Shen, Wei and Xie, Cihang and Yuille, Alan and Kong, Tao},
  journal={arXiv preprint arXiv:2111.07832},
  year={2021}
}

@article{mambalinearattention,
  title={Demystify Mamba in Vision: A Linear Attention Perspective},
  author={Han, Dongchen and Wang, Ziyi and Xia, Zhuofan and Han, Yizeng and Pu, Yifan and Ge, Chunjiang and Song, Jun and Song, Shiji and Zheng, Bo and Huang, Gao},
  journal=nips,
  year={2025}
}

@article{distillation,
  title={Distilling the knowledge in a neural network},
  author={Hinton, Geoffrey and Vinyals, Oriol and Dean, Jeff},
  journal={arXiv preprint arXiv:1503.02531},
  year={2015}
}

@article{mambaout,
  title={Mambaout: Do we really need mamba for vision?},
  author={Yu, Weihao and Wang, Xinchao},
  journal={arXiv preprint arXiv:2405.07992},
  year={2024}
}

@article{patchscale,
  title={Scaling Laws in Patchification: An Image Is Worth 50,176 Tokens And More},
  author={Wang, Feng and Yu, Yaodong and Wei, Guoyizhe and Shao, Wei and Zhou, Yuyin and Yuille, Alan and Xie, Cihang},
  journal={arXiv preprint arXiv:2502.03738},
  year={2025}
}

@article{linformer,
  title={Linformer: Self-attention with linear complexity},
  author={Wang, Sinong and Li, Belinda Z and Khabsa, Madian and Fang, Han and Ma, Hao},
  journal={arXiv preprint arXiv:2006.04768},
  year={2020}
}

@inproceedings{linattn,
  title={Transformers are rnns: Fast autoregressive transformers with linear attention},
  author={Katharopoulos, Angelos and Vyas, Apoorv and Pappas, Nikolaos and Fleuret, Fran{\c{c}}ois},
  booktitle=icml,
  year={2020}
}

@article{performer,
  title={Rethinking attention with performers},
  author={Choromanski, Krzysztof and Likhosherstov, Valerii and Dohan, David and Song, Xingyou and Gane, Andreea and Sarlos, Tamas and Hawkins, Peter and Davis, Jared and Mohiuddin, Afroz and Kaiser, Lukasz and others},
  journal={arXiv preprint arXiv:2009.14794},
  year={2020}
}

@inproceedings{nystromformer,
  title={Nystr{\"o}mformer: A nystr{\"o}m-based algorithm for approximating self-attention},
  author={Xiong, Yunyang and Zeng, Zhanpeng and Chakraborty, Rudrasis and Tan, Mingxing and Fung, Glenn and Li, Yin and Singh, Vikas},
  booktitle=aaai,
  year={2021}
}

@article{lstm,
  title={Long short-term memory},
  author={Hochreiter, Sepp and Schmidhuber, J{\"u}rgen},
  journal={Neural computation},
  year={1997}
}

@article{lstm01,
  title={Learning to forget: Continual prediction with LSTM},
  author={Gers, Felix A and Schmidhuber, J{\"u}rgen and Cummins, Fred},
  journal={Neural computation},
  year={2000}
}

@article{dim,
  title={Dim: Diffusion mamba for efficient high-resolution image synthesis},
  author={Teng, Yao and Wu, Yue and Shi, Han and Ning, Xuefei and Dai, Guohao and Wang, Yu and Li, Zhenguo and Liu, Xihui},
  journal={arXiv preprint arXiv:2405.14224},
  year={2024}
}

@inproceedings{zigma,
  title={Zigma: A dit-style zigzag mamba diffusion model},
  author={Hu, Vincent Tao and Baumann, Stefan Andreas and Gui, Ming and Grebenkova, Olga and Ma, Pingchuan and Fischer, Johannes and Ommer, Bj{\"o}rn},
  booktitle=eccv,
  year={2024}
}

@article{dig,
  title={Dig: Scalable and efficient diffusion models with gated linear attention},
  author={Zhu, Lianghui and Huang, Zilong and Liao, Bencheng and Liew, Jun Hao and Yan, Hanshu and Feng, Jiashi and Wang, Xinggang},
  journal={arXiv preprint arXiv:2405.18428},
  year={2024}
}

@article{vlmamba,
  title={Vl-mamba: Exploring state space models for multimodal learning},
  author={Qiao, Yanyuan and Yu, Zheng and Guo, Longteng and Chen, Sihan and Zhao, Zijia and Sun, Mingzhen and Wu, Qi and Liu, Jing},
  journal={arXiv preprint arXiv:2403.13600},
  year={2024}
}

@article{cobra,
  title={Cobra: Extending mamba to multi-modal large language model for efficient inference},
  author={Zhao, Han and Zhang, Min and Zhao, Wei and Ding, Pengxiang and Huang, Siteng and Wang, Donglin},
  journal={arXiv preprint arXiv:2403.14520},
  year={2024}
}

@article{kasai2021finetuning,
  title={Finetuning pretrained transformers into rnns},
  author={Kasai, Jungo and Peng, Hao and Zhang, Yizhe and Yogatama, Dani and Ilharco, Gabriel and Pappas, Nikolaos and Mao, Yi and Chen, Weizhu and Smith, Noah A},
  journal={arXiv preprint arXiv:2103.13076},
  year={2021}
}

@article{zhang2024hedgehog,
  title={The hedgehog \& the porcupine: Expressive linear attentions with softmax mimicry},
  author={Zhang, Michael and Bhatia, Kush and Kumbong, Hermann and R{\'e}, Christopher},
  journal={arXiv preprint arXiv:2402.04347},
  year={2024}
}

@article{mercat2024linearizing,
  title={Linearizing large language models},
  author={Mercat, Jean and Vasiljevic, Igor and Keh, Sedrick and Arora, Kushal and Dave, Achal and Gaidon, Adrien and Kollar, Thomas},
  journal={arXiv preprint arXiv:2405.06640},
  year={2024}
}

@article{wang2024mamba,
  title={The mamba in the llama: Distilling and accelerating hybrid models},
  author={Wang, Junxiong and Paliotta, Daniele and May, Avner and Rush, Alexander and Dao, Tri},
  journal=nips,
  year={2024}
}

@article{zhang2024lolcats,
  title={LoLCATs: On Low-Rank Linearizing of Large Language Models},
  author={Zhang, Michael and Arora, Simran and Chalamala, Rahul and Wu, Alan and Spector, Benjamin and Singhal, Aaryan and Ramesh, Krithik and R{\'e}, Christopher},
  journal={arXiv preprint arXiv:2410.10254},
  year={2024}
}

@article{liao2025multimodal,
  title={Multimodal Mamba: Decoder-only Multimodal State Space Model via Quadratic to Linear Distillation},
  author={Liao, Bencheng and Tao, Hongyuan and Zhang, Qian and Cheng, Tianheng and Li, Yingyue and Yin, Haoran and Liu, Wenyu and Wang, Xinggang},
  journal={arXiv preprint arXiv:2502.13145},
  year={2025}
}

@article{mvar,
  title={M-var: Decoupled scale-wise autoregressive modeling for high-quality image generation},
  author={Ren, Sucheng and Yu, Yaodong and Ruiz, Nataniel and Wang, Feng and Yuille, Alan and Xie, Cihang},
  journal={arXiv preprint arXiv:2411.10433},
  year={2024}
}

@inproceedings{flowar,
  title={FlowAR: Scale-wise Autoregressive Image Generation Meets Flow Matching},
  author={Ren, Sucheng and Yu, Qihang and He, Ju and Shen, Xiaohui and Yuille, Alan and Chen, Liang-Chieh},
  booktitle = {ICML},
  year = {2025}
}
